\documentclass[journal]{IEEEtran}
\usepackage[dvips]{graphicx}
\DeclareGraphicsExtensions{.pdf,.jpeg,.png}
\usepackage{multirow}
\usepackage{amsthm}
\usepackage{amsmath,amssymb,bm}
\usepackage{amsfonts,dsfont,color,bbm}
\usepackage{algorithm}
\usepackage{algorithmic}
\usepackage[font=footnotesize,labelfont=bf,justification=justified,singlelinecheck=false]{caption}
\usepackage{subfigure}
\usepackage{cite}
\usepackage{url}
\usepackage{cleveref}

\usepackage[utf8]{inputenc}

\crefname{equation}{}{}
\newtheorem{remark}{Remark}

\usepackage{setspace}

\newcommand{\be}{\begin{equation}}
\newcommand{\ee}{\end{equation}}
\newcommand{\bea}{\begin{eqnarray}}
\newcommand{\eea}{\end{eqnarray}}

\newcommand{\Real}{{\mathbbm{R}}}
\newcommand{\Lagr}{\mathcal{L}}

\AtBeginDocument{
\addtolength{\abovedisplayskip}{-0.5ex}
\addtolength{\abovedisplayshortskip}{-0.4ex}
\addtolength{\belowdisplayskip}{-0.5ex}
\addtolength{\belowdisplayshortskip}{-0.4ex}
\addtolength{\belowcaptionskip}{-3.6ex}
}

\begin{document}

\title{Unsupervised Online Anomaly Detection On Irregularly Sampled Or Missing Valued Time-Series Data Using LSTM Networks}

\author{Oguzhan Karaahmetoglu, Fatih Ilhan, Ismail Balaban and Suleyman S. Kozat \textit{Senior Member, IEEE}\thanks{This work is supported in part by Outstanding Researcher Programme Turkish Academy of Sciences and also in part by Tübitak Contract No: 117E153.}
\thanks{O. Karaahmetoglu, F. Ilhan and S. S. Kozat are with the Department of Electrical and Electronics Engineering, Bilkent University, Bilkent, Ankara 06800, Turkey, Tel: +90 (312) 290-2336, Fax: +90 (312) 290-1223, {(contact e-mail: \{koguzhan, filhan, kozat\}@ee.bilkent.edu.tr).}}
\thanks{O. Karaahmetoglu, F. Ilhan, I. Balaban and S. S. Kozat are also with the DataBoss A.S., Bilkent Cyberpark, Ankara 06800, (email: \{oguzhan.karaahmetoglu, fatih.ilhan, ismail.balaban, serdar.kozat\}@data-boss.com.tr)} }

\maketitle
\begin{abstract}
        We study anomaly detection and introduce an algorithm that processes variable length, irregularly sampled sequences or sequences with missing values. Our algorithm is fully unsupervised, however, can be readily extended to supervised or semi-supervised cases when the anomaly labels are present as remarked throughout the paper. Our approach uses the Long Short Term Memory (LSTM) networks in order to extract temporal features and find the most relevant feature vectors for anomaly detection. We incorporate the sampling time information to our model by modulating the standard LSTM model with time modulation gates. After obtaining the most relevant features from the LSTM, we label the sequences using a Support Vector Data Descriptor (SVDD) model. We introduce a loss function and then jointly optimize the feature extraction and sequence processing mechanisms in an end-to-end manner. Through this joint optimization, the LSTM extracts the most relevant features for anomaly detection later to be used in the SVDD, hence completely removes the need for feature selection by "expert" knowledge. Furthermore, we provide a training algorithm for the online setup, where we optimize our model parameters with individual sequences as the new data arrives. Finally, on real-life datasets, we show that our model significantly outperforms the standard approaches thanks to its combination of LSTM with SVDD and joint optimization. 

\end{abstract}

\section{Introduction}\label{sec:introduction}

\subsection {Preliminaries}
    Anomaly detection is the problem of detecting anomalous points or sequences, which models a wide range of real-life applications such as fault detection \cite{kiciman2005detecting}, fraud detection \cite{8038008}, surveillance \cite{li2013anomaly} and intrusion detection \cite{ryan1998intrusion}. In certain applications, a sequence processing system is required, which aims to detect anomalous sequences \cite{9064715}, \cite{7880607}. In such applications, anomalies are detected from the temporal context rather than the individual samples as the samples are auto-correlated and context information is needed, e.g. deviation from a normal pattern \cite{chandola2010anomaly}. However, in most applications, acquiring annotated data may not be possible due to high labeling cost such as in network anomaly detection or even impossible, e.g. zero-day attacks \cite{aleroud2012contextual}, \cite{7412753}, which restricts using labeled data \cite{zhang2006anomaly}, \cite{8267488}. In addition to the problems due to the lack of annotated data, the sequences can be irregularly sampled or have variable lengths \cite{li2016scalable}, \cite{lipton2015learning}. Here, we study the anomaly detection problem in an unsupervised framework, where we observe irregularly sampled, variable length sequences or sequences with missing values and introduce a system based on LSTM networks to extract temporal features and the SVDD classifier to classify the sequences without requiring explicit annotations for training.

	Due to its critical applications, various solutions have been proposed for this problem. One-Class Support Vector Machine (OC-SVM) \cite{scholkopf2001estimating} and the Support Vector Data Descriptors (SVDD) \cite{scholkopf2001estimating} models are two of the most popular approaches among the solutions \cite{erfani2016high}, \cite{8052229}. Both the OC-SVM and the SVDD are classification models that can separate outlier points via estimating the support of the data without requiring explicit annotations of the instances. Despite their advantage on detecting point anomalies, these classifiers have no inherent memory to model temporal patterns in the sequences. For fixed length sequences, the whole sequence is fed to the classifier in certain approaches \cite{wang2004anomaly}, \cite{zhang2007one}. For the variable length sequences, a sliding window is used to capture temporal dependencies \cite{ma2003time}. However, the performance of this approach is highly affected by the window length selection \cite{ma2003time}. Furthermore, these approaches lack or do not consider the time information when the sequences are irregularly sampled or have missing values, which is usually the case in real-life anomaly detection applications \cite{liao2005clustering}, \cite{vio2000reconstruction}.
	
	To solve the problems regarding the temporal pattern extraction, Recurrent Neural Networks (RNN)  \cite{goller1996learning} are used \cite{8438512}. RNN and its variants are capable of capturing temporal dependencies on sequences thanks to their inherent memory \cite{goller1996learning}. Moreover, certain types of RNNs such as the Gated Recurrent Unit (GRU) \cite{dey2017gate} and the Long Short Term Memory (LSTM) \cite{hochreiter1997long} have an internal gated structure, which allows them to be trained on long sequences by preventing the exploding and vanishing gradient problems \cite{hochreiter1997long}. Despite their superior ability in capturing the temporal patterns, these models accept regularly sampled instances. To this end, various solutions have been proposed. One of those approaches use an exponential decaying mechanism to formulate the definition of the state vector at any time instead of fixed sampling times \cite{mei2017neural}. Another approach uses LSTM model with time modulation gates \cite{sahin2018nonuniformly}. 
	
	In this paper, we introduce an anomaly detection model that processes irregularly sampled, variable length sequences or sequences with missing values in an unsupervised manner. This approach can be readily extended  to supervised or semi-supervised cases when the labels are available as presented remarks in the paper. Our approach is based on the LSTM network, which has superior abilities on time series data as it avoids the vanishing gradient problem \cite{hochreiter1997long}. Since we are processing irregularly sampled sequences or sequences with missing values due to application specific problems, we introduce a novel LSTM architecture with modulation gates particularly introduced for anomaly detection. Moreover, for the classification part, we use the SVDD to declare anomalies using the temporal features that are extracted by the LSTM model, which eliminates the need for feature selection, where the full system is optimized end-to-end. Although we use an SVDD in the final stage, other methods can be readily used as provided remarks in the paper. We jointly optimize both the feature extraction, i.e. the LSTM and the one-class classifier model. Finally, we also provide a training algorithm for online setup where we optimize our model with individual sequences with simple gradient-based updates.

\subsection {Prior Art and Comparisons}
    One-class classifiers have been applied to various anomaly detection tasks \cite{erfani2016high}, \cite{8052229}. The main advantage of the one-class classifiers is that they do not require any annotation for the input data and they can estimate the support of the data. Although the support estimation is a simple structure such as a hyperplane or a hypersphere, with the use of feature transformation functions, the data can be transformed to an infinite-dimensional, linearly separable space to boost the performance \cite{scholkopf2001estimating}, \cite{scholkopf2001kernel}. Despite the performance of one-class classifiers on detecting point anomalies, their performance on detecting contextual anomalies on time-series data is limited as they do not consider the context \cite{chandola2010anomaly}. In such cases, a fixed length time window is used to extend the standard classifier for time-series processing \cite{zhang2007one}, \cite{ma2003time}. However, using a fixed length time window to capture temporal dependencies highly affects the performance with respect to the chosen window size \cite{ma2003time}. Moreover, these methods fail to capture timing information for the irregularly sampled sequences or sequences with missing values \cite{vio2000reconstruction}. 
    
    Deep neural networks have shown impressive performance in various real-life problems as they are able to capture non-trivial patterns in data \cite{goller1996learning}, \cite{hochreiter1997long}. Certain deep models are specialized for particular tasks such as the RNNs, which have been extensively used on time-series data thanks to their inherent memory. Instead of using a fixed length time window, an RNN is used to capture temporal patterns with a state vector \cite{graves2013speech}, \cite{ergen2017unsupervised}. A specific variant of RNNs, the LSTM, has internal gates such as forget, input and output gates, which update the state vector with each new input vector \cite{hochreiter1997long}. This gated structure allows LSTM models to be used on longer sequences compared to the standard RNN by preventing the exploding and vanishing gradient problems \cite{hochreiter1997long}. Although RNNs have been successful in many problems, these deep models accept structured data for optimization and prediction, i.e. the input and output data have a fixed size. For time-series inputs on RNNs, each instance should be sampled regularly with a fixed sampling period \cite{goller1996learning}, \cite{hochreiter1997long}, which limits the application of these deep models on important real-life anomaly detection tasks \cite{sahin2018nonuniformly}, \cite{neil2016phased}. In certain approaches, the sampling time information is incorporated to the model via time modulation gates \cite{sahin2018nonuniformly}.

    To overcome the problems related to capturing long temporal patterns in sequences, we use an LSTM model to extract temporal features. After obtaining temporal features, we use a one-class classifier to label the sequences. Therefore, we do not require annotation for the training data and train our model in an unsupervised manner. Furthermore, thanks to the inherent memory of the LSTM model, the extracted temporal features are not limited with a fixed temporal length. Finally, to circumvent the issues caused by irregular sampling times, we use an LSTM model with time modulation gates for anomaly detection. Although the time modulation gates were used in time series regression \cite{sahin2018nonuniformly}, this idea is tailored to supervised framework whereas we are working in unsupervised framework. Unlike \cite{sahin2018nonuniformly}, we are able to optimize our model end-to-end in a fully unsupervised manner when there is no annotation for the input data, which is typical in various real-life anomaly detection problems \cite{zhang2006anomaly}, \cite{9059022}, \cite{8825555}. Unlike \cite{sahin2018nonuniformly}, we introduce a loss function, which is a combination of the reconstruction loss for the sequences and the one-class classifier loss. Through this loss function, we minimize the reconstruction loss while we optimize the support of the data, which prevents the trivial solutions of LSTM parameters when the whole model is trained with the classifier loss alone \cite{ergen2017unsupervised}.

\subsection {Contributions}
    Our main contributions are:
    
    \begin{enumerate}
        \item We introduce a novel unsupervised anomaly detection architecture that extracts temporal features from sequences and detect outliers among the sequences, where the underlying data can be irregularly sampled or has missing values.
        
        \item Our method jointly optimizes the feature extraction and anomaly detection end-to-end and can be readily extended to supervised and semi-supervised cases when the labels are present.
        
        \item Our approach is generic so that the feature extracting stage, i.e. the LSTM model can be replaced by any other variant of RNN such as the standard RNN and the GRU, and the anomaly detection part can be replaced by other differentiable unsupervised classifiers such as the OC-SVM.
        
        \item We introduce specific updates for the online setup where we update parameters using individual sequences as new data arrives.
        
        \item Through an extensive set of experiments, we show that our model significantly outperforms the standard approaches on real-life datasets that consist of irregularly sampled variable length sequences.
    \end{enumerate}{}

\subsection {Organization of the Paper}
    In the following section, we first describe the main problem. We then give a brief information about the model structures. In Section \ref{sec:model}, we explain the details of our overall algorithm. Furthermore, we introduce the loss function that we are minimizing together with the joint optimization procedure, which updates all parts of our model jointly. Finally, we extend our algorithm to online setup. In Section \ref{sec:experiments}, we explain the test setup for performance measurements. We also demonstrate our model performance on real-life datasets. We finalize the paper with remarks in the conclusion section, Section \ref{sec:conclusion}.

\section{Model and Problem Description}\label{sec:prob_desc}
    We denote the vectors with bold and lowercase letters, e.g. $ \textbf{x} = [x^{(i)}]_{i=1}^{I} $ is a vector with $ I $ elements. We denote the matrices with bold and uppercase letters, e.g. $ \textbf{W} $. For matrix indexing, we use the $ \textbf{W}_{i, j} $ notation, which denotes the element at the $ i $th row and the $ j $th column. Unless otherwise stated, all vectors are real, i.e. $ x^{(i)} \in \Real $, and column vectors. For vectors and matrices, the notation $ \textbf{x}^T $ denotes the ordinary transpose and the notation $ \langle \textbf{x}, \textbf{x} \rangle = ||\textbf{x}||^2 $ denotes the $ l^2 $ norm of the vector $ \textbf{x} $. Finally, $ \mathbbm{1} $ refers to a column vector of all ones where the length is understood from the context.

    \begin{figure}
        \centering
        \includegraphics[width=0.5\textwidth]{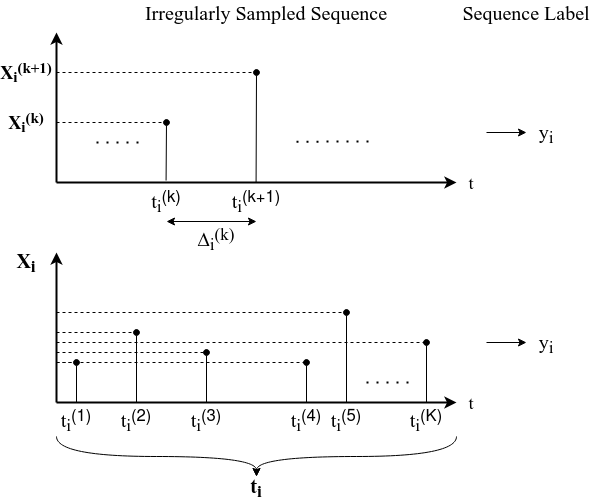}
        \caption{Illustration of an irregularly sampled sequence. Each sequence has a binary label, $ y_i $, indicating whether the sequence has nominal or anomalous behavior.}
        \label{fig:irregular_data}
    \end{figure}

	In the anomaly detection problem, we are given a set of temporal sequences $ \{\textbf{X}_i\}_{i} $ where each sequence can have a different length $ \textbf{X}_i = [\textbf{x}_i^{(k)}]_{k=1}^{K_i} $, where $ \textbf{x}_i^{(k)} \in \Real^M $. These sequences have corresponding binary labels $ \{y_i\}_{i} $ where $ y_i \in \{-1, +1\} $ that indicate the nominal and anomalous sequences, respectively. Such sequences are visualized in Fig. \ref{fig:irregular_data} for one dimensional $ x_i^{(k)} \in \Real $. We model these series as irregularly sampled time series data. Therefore, we define another set $ \{\textbf{t}_i\}_{i} $ that contains the sampling times of every element $ \textbf{t}_i = [t_i^{(k)}]_{k=1}^{K_i} $, for all sequences, which is ordered in time, i.e. $ t_i^{(k_1)} > t_i^{(k_2)} $ for $ k_1 > k_2 $ as in Fig. \ref{fig:irregular_data}. Thus, we model $ \Delta_i^{(k)} \neq \Delta_i^{(l)} $ for $ k \neq l $ where $ \Delta_i^{(k)} = t_i^{(k)} - t_i^{(k-1)} $ for $ k \in \{2, ..., K_i\} $. An example irregularly sampled sequence is also shown in Fig. \ref{fig:irregular_data}. Although we illustrate the sequence with 1-$ D $ samples, i.e. $ \textbf{x}_i^{(k)} \in \Real $, our formulation covers any $ M $-$ D $ vectors.
	
	As in Fig. \ref{fig:irregular_data}, sequences have binary labels $ y_i \in \{-1,\ +1\} $. Our aim is to classify a sequence $ \textbf{X}_i $ with the sampling times $ \textbf{t}_i $ as a nominal sequence, $ \hat{y}_i = -1 $, or as an anomaly sequence, $ \hat{y}_i = +1 $. Hence, we introduce a function
	
	\begin{equation}
	    \hat{y}_i = F(\textbf{X}_i, \textbf{t}_i;\boldsymbol{\theta}),
	\end{equation}
	which labels the sequences as nominal or anomalous. We optimize the function by minimizing a loss function with respect to the set of function parameters $ \boldsymbol{\theta} $. This loss function has the form
	
	\begin{equation}
	    L(X, T) = \sum_{i=1}^I l(\textbf{X}_i, \textbf{t}_i) + L_{\text{reg}},
	\end{equation}
	where $ l(\textbf{X}_i, \textbf{t}_i) $ is the individual loss contribution from the $ i $th sequence in the dataset and the term $ L_{\text{reg}} $ is the regularization term independent of the sequences. For the batch setup, we use a known $ I $ and for the online setup, we choose $ I $ as 1. Structures and the details of these components will be provided in the next section.
	
	\begin{figure}
	    \centering
	    \includegraphics[width=0.5\textwidth]{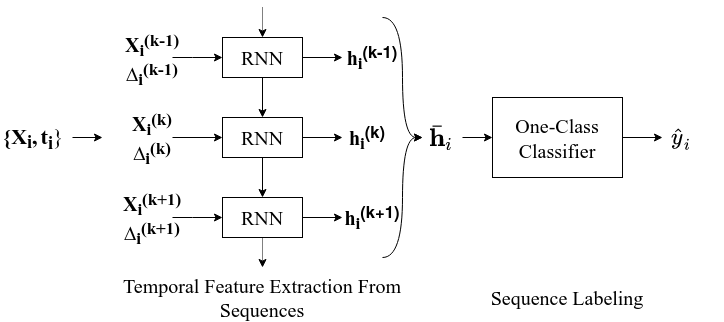}
	    \caption{Overall diagram of the introduced model. For a sequence with samples $ \textbf{X}_i $ and sampling times $ \textbf{t}_i $, a feature vector is generated by feeding the samples sequentially. Using the feature vector, sequences are labeled via a one-class classifier.}
	    \label{fig:model_structure}
	\end{figure}{}
	
	We model the function $ F $ as two subblocks, an RNN or LSTM model that extracts temporal features, which generates a representation of the sequence, and a classifier model that classifies these representations as nominal or anomalous. This structure is depicted in Fig. \ref{fig:model_structure}. First, the sequence $ \textbf{X}_i $ and its sampling time sequence $ \textbf{t}_i $ is fed to the RNN model. RNN model generates a fixed length representation $ \bar{\textbf{h}}_i $ from the variable length sequences. This representation is used to classify the sequence. We describe the mechanisms to adapt the RNN model to irregularly sampled variable length sequences in the following subsection.

	\subsection{Recurrent Networks on Irregularly Sampled Series}
	    Here, we describe the approaches based on the standard RNN structure to accept irregularly sampled time series data. In the standard approach, RNNs are designed to work with regularly sampled time-series data \cite{goller1996learning}. For a sequence of regularly sampled inputs $ \textbf{X}_i = [\textbf{x}_i^{(k)}]_{k=1}^{K_i} $, RNN state equation is given as 
	    
	    \begin{equation}
	    \begin{split}
	        \textbf{h}_i^{(k)} &= f(\textbf{W}_\textbf{x}\textbf{x}_i^{(k)} + \textbf{W}_\textbf{h}\textbf{h}_i^{(k-1)} + \textbf{b}).
	    \end{split}{}
	    \label{eqn:rnn_eqs}
	    \end{equation}
The vector $ \textbf{h}_i^{(k)} $ is the state vector of the RNN model, which is the internal memory storing temporal features extracted from the sequence $ \textbf{X}_i $. Other variants of the RNNs have different state equations. In our model, we specifically use the LSTM model, which has the set of equations,

	    \begin{equation}
	        \begin{split}
	            \textbf{c}_i^{(k)} &= \textbf{c}_i^{(k-1)} \odot \textbf{f}_i^{(k)} + \textbf{g}_i^{(k)} \odot \textbf{i}_i^{(k)},\\
	            \textbf{h}_i^{(k)} &= g(\textbf{c}_i^{(k)}) \odot \textbf{o}_i^{(k)},
	            \label{eqn:lstm_eqs}
	        \end{split}{}
	    \end{equation}
where
	    
	    \begin{equation}
	    \begin{split}
	        \textbf{f}_i^{(k)} &= \sigma(\textbf{W}_{\textbf{fx}} \textbf{x}_i^{(k)} + \textbf{W}_{\textbf{fh}} \textbf{h}_i^{(k-1)} + \textbf{b}_\text{f}),\\
	        \textbf{i}_i^{(k)} &= \sigma(\textbf{W}_{\textbf{ix}} \textbf{x}_i^{(k)} + \textbf{W}_{\textbf{ih}} \textbf{h}_i^{(k-1)} + \textbf{b}_\text{i}),\\
	        \textbf{g}_i^{(k)} &= g(\textbf{W}_{\textbf{gx}} \textbf{x}_i^{(k)} + \textbf{W}_{\textbf{gh}} \textbf{h}_i^{(k-1)} + \textbf{b}_\text{g}),\\
	        \textbf{o}_i^{(k)} &= \sigma(\textbf{W}_{\textbf{ox}} \textbf{x}_i^{(k)} + \textbf{W}_{\textbf{oh}} \textbf{h}_i^{(k-1)} + \textbf{b}_\text{o}).\\
	    \end{split}{}
	    \label{eqn:lstm_gates}
	    \end{equation}
Matrices $ \{ \textbf{W}_{\textbf{fx}}, \textbf{W}_{\textbf{ix}}, \textbf{W}_{\textbf{gx}}, \textbf{W}_{\textbf{ox}} \} $ and $ \{ \textbf{W}_{\textbf{fh}}, \textbf{W}_{\textbf{ih}}, \textbf{W}_{\textbf{gh}}, \textbf{W}_{\textbf{oh}} \} $ are the weight matrices and $ \{ \textbf{b}_{\text{f}}, \textbf{b}_{\text{i}}, \textbf{b}_{\text{g}}, \textbf{b}_{\text{o}} \} $ are the bias vectors. Functions $ \sigma $ and $ g $ are nonlinear activation functions, which are generally chosen as the Sigmoid ($ \sigma(x) = 1 / (1 + e^{-x}) $) and the Tanh ($ \text{tanh}(x) = (e^x - e^{-x})/(e^x + e^{-x}) $) function respectively.
	    
	    As in \eqref{eqn:lstm_gates}, the LSTM model has a gated structure, which performs different tasks such as forgetting the state vector, updating the state vector with the current input. This gated structure prevents the vanishing gradients problem especially for long sequences, thus allowing the model to capture longer temporal dependencies \cite{hochreiter1997long}.
	    
    \begin{figure*}
        \centering
        \includegraphics[width=0.8\textwidth]{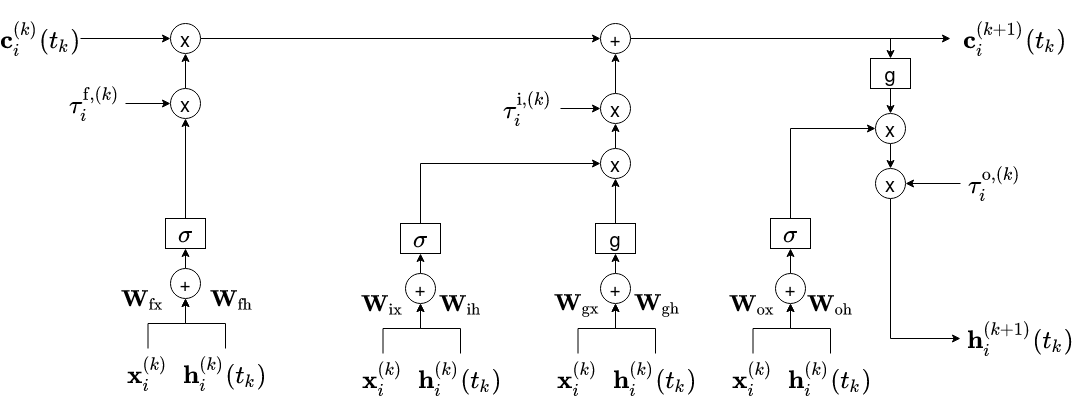}
        \caption{LSTM architecture with additional time modulation gates. Single update for a sequence is shown. In addition to the standard LSTM gates, additional time modulation gates are used to modulate the outputs of the internal gates.}
        \label{fig:tglstm}
    \end{figure*}    
	    
	    For every element in the sequence, the state vector is updated at regular times. In irregularly sampled series, the sampling time of the elements contain information as the sampling period is not fixed. There are certain improvements on the RNNs such as incorporating the sampling time, $ \Delta_i^{(k)} $, as an additional feature in the input vector as
	    
	    \begin{equation}
	    	 \textbf{h}_i^{(k)} = f(\textbf{W}_\textbf{x}\textbf{x}_i^{(k)} + \textbf{W}_\textbf{h}\textbf{h}_i^{(k-1)} + \textbf{W}_\textbf{t} \Delta_i^{(k)} + \textbf{b})
	    	 \label{eqn:rnn_additive}
	    \end{equation}
	    for the standard RNN state transition. However, in this approach, the sampling time information has an additive effect on the state transition, which may not be suitable in most cases \cite{sahin2018nonuniformly}. In a different approach, state decaying with an exponential kernel \cite{mei2017neural} is used. In the state decaying mechanism, the state vector is updated at irregular sampling times. The state vector $ \textbf{h}_i^{(k)} $ in \eqref{eqn:lstm_eqs} is formulated at any time as
	    
	    \begin{equation}
	        \textbf{h}_i^{(k)}(t) = \textbf{h}_i^{(k)} e^{-\gamma(t - t_k)},
	        \label{eqn:state_decay}
	    \end{equation}
where $ \textbf{h}_i^{(k)} $ is the state updated with the $ k $th element of the sequence, $ \textbf{x}_i^{(k)} $. The parameter $ \gamma $ is the decay rate, which controls the long term behavior of the state vector. We use the notation $ t_k $ instead of $ t_i^{(k)} $ for notational simplicity. We then reformulate the state update equations at the time of the $ k $th element as,
	    
	   \begin{equation}
	   \begin{split}
	       \textbf{c}_i^{(k)}(t_k) &= \textbf{c}_i^{(k-1)}(t_{k}) \odot \textbf{f}_i^{(k)}(t_k) + \textbf{g}_i^{(k)}(t_k) \odot \textbf{i}_i^{(k)}(t_k),\\
	       \textbf{h}_i^{(k)}(t_k) &= g(\textbf{c}_i^{(k)}(t_k)) \odot \textbf{o}_i^{(k)}(t_k),
	   \end{split}{}
	   \label{eqn:state_decay_update}
	   \end{equation}{}
	   where
	   
	   \begin{equation}
	       \begin{split}
	        \textbf{f}_i^{(k)}(t) &= \sigma(\textbf{W}_{\textbf{fx}} \textbf{x}_i^{(k)} + \textbf{W}_{\textbf{fh}} \textbf{h}_i^{(k-1)}(t) + \textbf{b}_\text{f}),\\
	        \textbf{i}_i^{(k)}(t) &= \sigma(\textbf{W}_{\textbf{ix}} \textbf{x}_i^{(k)} + \textbf{W}_{\textbf{ih}} \textbf{h}_i^{(k-1)}(t) + \textbf{b}_\text{i}),\\
	        \textbf{g}_i^{(k)}(t) &= g(\textbf{W}_{\textbf{gx}} \textbf{x}_i^{(k)} + \textbf{W}_{\textbf{gh}} \textbf{h}_i^{(k-1)}(t) + \textbf{b}_\text{g}),\\
	        \textbf{o}_i^{(k)}(t) &= \sigma(\textbf{W}_{\textbf{ox}} \textbf{x}_i^{(k)} + \textbf{W}_{\textbf{oh}} \textbf{h}_i^{(k-1)}(t) + \textbf{b}_\text{o}).\\
	       \end{split}{}
	   \end{equation}
	   for $ t > t_k $. Thus, we can update the state of the model at any timestep for the given sequence. However, note that the performance of such approaches heavily depend on the decay rate and do not fully exploit the true nature of the irregularity in the data \cite{sahin2018nonuniformly}.

	   Another approach is learning a modulation gate \cite{sahin2018nonuniformly}. In this approach, in addition to the gates of the LSTM model, additional time modulation gates are added to each gate as shown in Fig. \ref{fig:tglstm}. These modulated gates differ from the standard LSTM equations as
	    
	    \begin{equation}
	        \begin{split}
	            \textbf{c}_i^{(k)} &= \textbf{c}_i^{(k-1)} \odot \textbf{f}_i^{(k)} \odot \tau_{i}^{\text{f},\ (k)} + \textbf{g}_i^{(k)} \odot \textbf{i}_i^{(k)} \odot \tau_{i}^{\text{i},\ (k)}\\
	            \textbf{h}_i^{(k)} &= g(\textbf{c}_i^{(k)}) \odot \textbf{o}_i^{(k)} \odot \tau_{i}^{\text{o},\ (k)}
	        \end{split}{}
	        \label{eqn:gated_lstm}
	    \end{equation}{}
	    where,
	    
	    \begin{equation}
	        \begin{split}
	            \tau_{i}^{\text{f},\ (k)} &= f(\textbf{W}_{\text{ft}} \mathbf{\Delta}_i^{(k)}),\\
	            \tau_{i}^{\text{i},\ (k)} &= f(\textbf{W}_{\text{it}} \mathbf{\Delta}_i^{(k)}),\\
	            \tau_{i}^{\text{o},\ (k)} &= f(\textbf{W}_{\text{ot}} \mathbf{\Delta}_i^{(k)}).\\
	        \end{split}
	    \end{equation}
	    Here the vectors $ \tau_{i}^{\text{f},\ (k)} $, $ \tau_{i}^{\text{i},\ (k)} $ and $ \tau_{i}^{\text{o},\ (k)} $ are the time modulation gates that are multiplied with each gate except the input gate. $ \mathbf{\Delta}_i^{(k)} $ is a vector of any nonlinear functions of the last sampling duration $ \Delta_i^{(k)} $.
	    
	    \begin{remark}
	        It is possible to reformulate the standard RNN or GRU structure to directly embed the modulation gates to the model. For the standard RNN, we formulate the update equation as
	        
	        \begin{equation*}
	        \begin{split}
	            &\textbf{h}_i^{(k)}(t_i^{(k)}) = f(\textbf{W}_\textbf{\emph{x}}\textbf{x}_i^{(k)} + \textbf{W}_\textbf{\emph{h}}\textbf{h}_i^{(k-1)}(t_k) + \textbf{b}) \odot \mathbf{\tau}_i^{(k)},\\
	         	&\mathbf{\tau}_i^{(k)} = f(\textbf{W}_\mathbf{\tau}\mathbf{\Delta}_i^{(k)}).
	        \end{split}
	        \label{eqn:rnn_update}
	        \end{equation*}
	        For the GRU, the update equation becomes
	        
	        \begin{equation*}
	        \begin{split}
	        &\textbf{z}_i^{(k)}(t) = \sigma(\textbf{W}_{\textbf{\emph{zx}}} \textbf{x}_i^{(k)} + \textbf{W}_{\textbf{\emph{zh}}} \textbf{h}_i^{(k-1)} + \textbf{b}_{\textbf{\emph{z}}}),\\
	        &\textbf{r}_i^{(k)}(t) = \sigma(\textbf{W}_{\textbf{\emph{rx}}} \textbf{x}_i^{(k)} + \textbf{W}_{\textbf{\emph{rh}}} \textbf{h}_i^{(k-1)} + \textbf{b}_{\textbf{\emph{r}}}),\\
	        &\hat{\textbf{h}}_i^{(k)}(t) = g(\textbf{W}_\textbf{\emph{hh}} (\textbf{r}_i^{(k)}(t) \odot \textbf{h}_i^{(k-1)}(t_k)) + \textbf{W}_\textbf{\emph{hx}} \textbf{x}_i^{(k)}), \\
	        &\textbf{h}_i^{(k)}(t_k) = (1 - \textbf{z}_i^{(k)}(t_k)) \odot \textbf{h}_i^{(k-1)}(t_{k}) \odot \mathbf{\tau}_i^{\text{\emph{h}},(k)} \\ 
	        &\ \ \ \ \ \ \ \ \ \ + \textbf{z}_i^{(k)}(t_k) \odot \hat{\textbf{h}}_i^{(k)}(t_k) \odot \mathbf{\tau}_i^{\emph{i},(k)}
	        \end{split}
	            \label{eqn:gru_update}
	        \end{equation*}
	        where
	        
	        \begin{equation}
	        	\begin{split}
	        	&\mathbf{\tau}_i^{\emph{h},(k)} = f(\textbf{W}_\mathbf{\emph{h}}\mathbf{\Delta}_i^{(k)}),\\
	        	&\mathbf{\tau}_i^{\emph{i},(k)} = f(\textbf{W}_\mathbf{\emph{i}}\mathbf{\Delta}_i^{(k)}).
	        	\end{split}
	        \end{equation}
	   \end{remark}

	We downscale the $ K_i $ state vectors by taking the state vector at the last time step, i.e. $ \textbf{h}_i^{(K)}(t_i^{(K_i)}) $.
	   
	   \begin{remark}
	        Instead of taking the state vector at the last step, which we refer to as the last pooling, we could also take the average of the state vectors through all sampling steps as
	        
	        \begin{equation}
	            \Bar{\textbf{h}}_i = \frac{1}{K}\sum_{k=1}^K \textbf{h}_i^{(k)}(t_i^{(k)}).
	        \end{equation}
	        We refer to this method as the mean pooling.
	   \end{remark}
	   For each sequence, we obtain a fixed length representation $ \Bar{\textbf{h}}_i $ using the last pooling or mean pooling method. In the following subsection, we describe the one-class unsupervised anomaly detection methods.

	  \subsection{Unsupervised Anomaly Detection with the One Class Classifiers}
	    Given a set of vectors $ \{\Bar{\textbf{h}}_i\}_{i=1}^I $, a one-class classifier labels each vector as nominal ($ y_i = -1 $) or anomaly ($ y_i = +1 $). Unlike the supervised methods, one-class classifiers do not require any annotated input for training as they estimate the support of the data.
	
        OC-SVM is a one-class adaptation of the SVM model, which estimates the support of the distribution with a hyperplane. Thus the decision function is
        
        \begin{equation}
            \hat{y}_i = \text{sign}(\textbf{W}\phi(\Bar{\textbf{h}}_i) - b),
            \label{eqn:ocsvm_decision}
        \end{equation}
        where $ \textbf{W} $ and $ b $ are the weight and bias vectors of the supporting hyperplane. The function $ \phi(\cdot) $ is the feature transformation function that maps the inputs to a linearly separable, infinite-dimensional space. To find the separating hyperplane, the optimization problem
        
        \begin{equation}
            \min \frac{||\textbf{W}||^2}{2} + \frac{1}{I\lambda} \sum_{i=1}^I \xi_i - b ,
        \label{eqn:ocsvm_opt}
        \end{equation}
        \begin{equation}
            \text{s.t. } \textbf{W}\phi(\Bar{\textbf{h}}_i) - b \geq -\xi_i,\ \xi_i \geq 0\ \forall i
            \label{eqn:ocsvm_subject}
        \end{equation}
        is solved, where $ \lambda $ is a regularization parameter and $ \xi_i $ are the slack variables penalizing the misclassified instances. The separating hyperplane is found using a quadratic programming based optimization \cite{scholkopf2001estimating}. However, it is also possible to find the hyperplane using a gradient based optimization by reformulating the slack variable in \eqref{eqn:ocsvm_opt} and \eqref{eqn:ocsvm_subject} using the hinge-loss as
        
        \begin{equation}
            \min \frac{||\textbf{W}||^2}{2} + \frac{1}{I\lambda} \sum_{i=1}^I \max (0, b - \textbf{W}\phi(\Bar{\textbf{h}}_i)) - b.
            \label{eqn:ocsvm_hinge}
        \end{equation}
        To make the hinge-loss in \eqref{eqn:ocsvm_hinge} differentiable, we smoothly approximate the max function with the softplus function as 
        
        \begin{equation}
            \min \frac{||\textbf{W}||^2}{2} + \frac{1}{I\lambda} \sum_{i=1}^I\psi(b - \textbf{W}\phi(\Bar{\textbf{h}}_i)) - b,
        \end{equation}
        \begin{equation}
            \psi(b - \textbf{W}\phi(\Bar{\textbf{h}}_i)) = \frac{1}{\beta} \log (1 + e^{\beta (b - \textbf{W}\phi(\Bar{\textbf{h}}_i)})),
            \label{eqn:}
        \end{equation}
        where log represents the natural logarithm and $ \beta $ is a smoothing parameter. Thus, the loss function for the hyperplane optimization is given by
        
        \begin{equation}
            \Lagr_\text{oc} = \frac{||\textbf{W}||^2}{2} + \frac{1}{I\lambda} \sum_{i=1}^I\psi(b - \textbf{W}\phi(\Bar{\textbf{h}}_i)) - b,
            \label{eqn:ocsvm_loss}
        \end{equation}
        which allows the optimization of $ \textbf{W} $ and $ b $ with the Stochastic Gradient Descent (SGD) method.
        
        The SVDD is also a one-class classifier and it estimates the support of the nominal instances with a hypersphere with radius $ r $ and center $ \textbf{c} $. Thus, the decision function is
        
        \begin{equation}
            \hat{y}_i = \text{sign}(||\phi(\Bar{\textbf{h}}_i) - \textbf{c}|| - r),
            \label{eqn:svdd_decision}
        \end{equation}
        where $ r > 0 $. Optimization problem for the hypersphere estimation is formulated as
        
		\begin{equation}
            \min r^2 + \frac{1}{I\lambda} \sum_{i=1}^I \xi_i,
        \label{eqn:svdd_opt}
        \end{equation}
        \begin{equation}
            \text{s.t. } ||\phi(\Bar{\textbf{h}}_i) - \textbf{c}||^2 - r^2 \leq \xi_i,\ \xi_i \geq 0\ \forall i.
            \label{eqn:svdd_subject}
        \end{equation}
        Consequently, the loss function for the SVDD is
        
        \begin{equation}
            \Lagr_\text{oc} = r^2 + \frac{1}{I\lambda} \sum_{i=1}^I\psi(||\phi(\Bar{\textbf{h}}_i) - \textbf{c}||^2 - r^2)
            \label{eqn:svdd_loss}
        \end{equation}{}
        Therefore, we can optimize the radius and the center of the supporting hypersphere with the SGD.

        \begin{remark}
        	The loss functions in \eqref{eqn:ocsvm_loss} and \eqref{eqn:svdd_loss} can be extended to the semi-supervised framework as
        	\begin{equation}
        	\begin{split}
            \Lagr_\text{oc} = \frac{||\textbf{W}||^2}{2} + \frac{1}{(I+J)\lambda} \sum_{i=1}^I\psi(y_i(b -  \textbf{W}\phi(\Bar{\textbf{h}}_i))) - b\\
            + \frac{\nu}{(I+J)\lambda} \sum_{j=1}^J\psi(b - \textbf{W}\phi(\hat{\textbf{h}}_j)) - b
            \end{split}
            \label{eqn:ocsvm_loss_semi}
        \end{equation}
        	for the OC-SVM and
        	
        	\begin{equation}
        	\begin{split}
            \Lagr_\text{oc} = r^2 + \frac{1}{(I+J)\lambda} \sum_{i=1}^I\psi(y_i(||\phi(\Bar{\textbf{h}}_i) - \textbf{c}||^2 - r^2))\\
             + \frac{\nu}{(I+J)\lambda} \sum_{j=1}^J\psi(||\phi(\hat{\textbf{h}}_i) - \textbf{c}||^2 - r^2)
            \end{split}
            \label{eqn:svdd_loss_semi}
            \end{equation}
        	for the SVDD, where $ \{\Bar{\textbf{h}}_i\}_{i=1}^{I} $ are the labeled instances and $ \{\hat{\textbf{h}}_j\}_{j=1}^{J} $ are the unlabeled instances and $ \nu \in \Real^+ $. It is possible to extend the formulations in \eqref{eqn:ocsvm_loss_semi} and \eqref{eqn:svdd_loss_semi} to the fully supervised setup by using only the sequence $ \{\hat{\textbf{h}}_j\}_{j=1}^{J} $.
        \end{remark}
        
        In the next section, we give the details of our overall anomaly detection algorithm.

\section{Anomaly Detection on Irregularly Sampled Sequences}\label{sec:model}
    Here, we describe the details of our unsupervised anomaly detection algorithm for irregularly sampled sequences, which also processes the sequences with missing values. We first describe the temporal feature extraction from sequences using the LSTM model with the updated gates described in the previous section. Later, we explain the sequence labeling mechanism from the temporal features using a one-class classifier.

    \subsection{Extracting Temporal Features From Irregularly Sampled Sequences}
        For a given input sequence $ \textbf{X}_i $ with the sampling times $ \textbf{t}_i $, we obtain the feature vector as shown in Fig. \ref{fig:model_structure}. We feed each element $ \textbf{x}_i^{(k)} $ and the sampling time differences $ \Delta_i^{(k)} $ sequentially and obtain a state vector $ \textbf{h}_i^{(k)} $ after each update. For this purpose, we use the LSTM model with the time modulation gates as it can model the irregularities in the sequence. Finally, we use a pooling method to obtain a fixed length feature vector $ \Bar{\textbf{h}}_i $ from the sequence $ \textbf{X}_i $.
        
        We optimize our feature vectors by minimizing the reconstruction loss
        
        \begin{equation}
            \Lagr_\text{recon} = \frac{1}{I}\sum_{i=1}^I \sum_{k=1}^K (f^D(\textbf{h}_i^{(k)}) - \textbf{x}_i^{(k)})^2
            \label{eqn:recon_loss}
        \end{equation}
        where $ \textbf{h}_i^{(0)} $ is a zero vector and we model $ f^D(\cdot) $ with a stack of $ D $ fully connected layers each with the relation
        
        \begin{equation}
            f(\textbf{x}) = g(\textbf{W}_d\textbf{x} + \textbf{b}_d),
            \label{eqn:dense_layer}
        \end{equation}
        where $ \textbf{W}_d $ and $ \textbf{b}_d $ are the weight and bias vectors for the layer $ d $ and $ g $ is a nonlinear activation function that we choose as the Rectified Linear Unit (ReLU, $ g(x) = \max(0, x) $). Hence, we train our LSTM structure so as to reconstruct the sequence at the irregular sampling times. Therefore, the LSTM model, together with the fully connected layers can be considered  as an autoencoder or prediction coder that encodes the characteristics of the sequences in the state vector.
        
         We choose the vector $ \mathbf{\Delta}_i^{k} $ in \eqref{eqn:gated_lstm} as the vector of powers of the sampling difference $ \Delta_i^{(k)} $, i.e. $ [{\Delta_i^{(k)}}^m]_{m=0}^\tau $. Our motivation for this vector and the reconstruction loss choice comes from the Taylor expansion of a sequence. A continuous sequence $ \textbf{X}(t) $ can be formulated as
         
         \begin{equation}
         	\textbf{X}(t) = \sum_{m=0}^{\infty} \textbf{X}^{(m)}(t_k) (t-t_k)^m,
         	\label{eqn:taylor}
         \end{equation}
         where $ \textbf{X}^{(i)}(t) $ is the $ i $th derivative of the sequence. Since we are given irregularly sampled sequences or sequences with missing values, we can represent a sample as
         \begin{equation}
         	\textbf{x}_i^{(k)} = \sum_{m=0}^{\tau} \textbf{f}_m(\textbf{h}_i^{(k-1)}) {\Delta_i^{(k)}}^m,
         	\label{eqn:taylor_seq}
         \end{equation}
         where we model the functions $ \{\textbf{f}_m\}_{m=0}^\tau $ with the LSTM structure with the modulation gates, which is a function of the state vector. Therefore, the LSTM structure and the time modulation gates are optimized to model the sequence. Note that neither the additive sampling time incorporation nor the state decaying mechanisms are capable of modeling such formulation.
        
        In the next subsection, we describe the labeling mechanism for our anomaly detection algorithm using the representations generated by the encoder LSTM model.

    \subsection{Anomaly Detection Using Feature Vectors}
        After extracting the temporal feature vectors from the sequences, we label the vectors $ \bar{\textbf{h}}_i $ as nominal or anomalous using the OC-SVM or SVDD method. Hence, our decision function has the form
        
        \begin{equation}
            \hat{y}_i = \text{sign}(\textbf{W}\phi(\bar{\textbf{h}}_i) - b)
            \label{eqn:ocsvm_decision}
        \end{equation}
        for the OC-SVM and
        
        \begin{equation}
            \hat{y}_i = \text{sign}(||\phi(\bar{\textbf{h}}_i) - \textbf{c}|| - r)
            \label{eqn:svdd_decision}
        \end{equation}
        for the SVDD. We convert the optimization problems in \eqref{eqn:ocsvm_opt} and \eqref{eqn:svdd_opt} as the loss functions in \eqref{eqn:ocsvm_loss} and \eqref{eqn:svdd_loss} and find the hyperplane/hypersphere parameters with the SGD. Note that for the softplus function approximation to the max function, we choose the smoothing parameter $ \beta = 100 $ and observe that it is sufficient for our experiments. Also, we have observed that the SVDD yields better results compared to the OC-SVM classifier in our experiments, hence we continue our description with the SVDD. Finally, we have observed that setting the feature transformation function $ \phi(\cdot) $ in \eqref{eqn:svdd_decision} to the identity function, i.e. $ \phi(x) = x $,  does not degrade the performance of our model.
        
        \begin{remark}
        Although the joint optimization of the LSTM and one-class classifier parts is desirable, directly optimizing the LSTM parameters using the hinge-loss would result in trivial solutions, i.e. the LSTM parameters will converge to zero values. To this end, we minimize the reconstruction loss while we optimize the support of the feature vectors. This is the main reason we introduce $ \Lagr_\emph{recon} $ unlike \cite{sahin2018nonuniformly}.
        \end{remark}
        
        In the following subsection, we describe the details of the joint optimization of our overall algorithm.

    \subsection{Joint Optimization of the Feature Extraction and the Outlier Detection}
    
        In this section, we first provide the loss function and the joint training procedure for our algorithm. Finally, we extend our training algorithm to online setup where we update the model parameters as we observe new sequences.
        
        Combining both the reconstruction loss in \eqref{eqn:ocsvm_loss} and the classifier loss in \eqref{eqn:recon_loss}, we obtain the weighted combination,
        
        \begin{equation}
            \Lagr = \Lagr_\text{oc} + \alpha \Lagr_\text{recon},
            \label{eqn:overall_loss}
        \end{equation}
        where $ \alpha $ is the weight parameter for the losses. Thanks to the reconstruction loss in the combined loss function, the LSTM parameters will be optimized so that the state vector encodes the necessary information to reconstruct the sequences and the features discriminating the anomalous inputs \cite{chen2018learning}. Although the RNNs were used as an unsupervised feature extracting stage in conjunction with regression and classification stages, thanks to the SVDD, our approach is able to remain fully unsupervised. We update the LSTM parameters and the classifier parameters jointly with the update
        
        \begin{equation}
        	\boldsymbol{\theta}_{n+1} = \boldsymbol{\theta}_n + \mu \frac{1}{B} \sum_{\textbf{x} \in X_B} \frac{\partial \Lagr}{\partial \boldsymbol{\theta}}(\textbf{x}),
        	\label{eqn:param_upd}
        \end{equation}
        where $ \theta_n $ is a model parameter (either an LSTM or SVDD parameter) at step $ n $ and $ X_B $ is a minibatch containing $ B $ randomly selected sequences. $ \mu $ is the learning rate, which controls the amount of parameter updates at each step. 
        
        \begin{algorithm}
        \caption{Anomaly detection training algorithm.}
        \begin{algorithmic}
        \REQUIRE  $ X_{\text{train}} $, $ T_{\text{train}} $
        \STATE Randomly initialize the LSTM model.
        \STATE Randomly initialize the One-Class Classifier
        \STATE $ n \leftarrow 1 $
        \WHILE{$ n < N_{\text{epoch}} + 1 $}
        	\STATE $ m \leftarrow 1 $
        	\WHILE{$ m < I_{\text{train}} / B + 1 $}
        		\STATE Randomly pick batch $ X_B = [\textbf{X}_b]_{b=1}^B$
            	\STATE $ \bar{H} \leftarrow [] $, $ \hat{X} \leftarrow [] $
            	\STATE $ b \leftarrow 1 $
            	\WHILE{$ b < B+1 $}
                	\STATE $ H_b \leftarrow [] $, $ \hat{X}_b \leftarrow [] $
                	\STATE $ k \leftarrow 1 $
                	\WHILE{$ k < K_b + 1 $}
                    	\STATE $ \textbf{h}_b^{(k)} \leftarrow \text{LSTM}(\textbf{x}_b^{(k)}) $
                    	\STATE $ H_b \leftarrow [H_b,\ \textbf{h}_b^{(k)}] $, $ \hat{X}_b \leftarrow [\hat{X}_b, f^D(\textbf{h}_b^{(k)})] $
                    	\STATE $ k \leftarrow k+1 $
                	\ENDWHILE
                	\STATE $ \bar{\textbf{h}}_b \leftarrow \text{Pool}(H_b) $
                	\STATE $ \bar{H} \leftarrow [\bar{H},\ \bar{\textbf{h}}_i] $, $ \hat{X} \leftarrow [\hat{X},\ \hat{X}_b] $
                	\STATE $ b \leftarrow b+1 $
        		\ENDWHILE
        	\STATE Compute $ \Lagr $ using \eqref{eqn:overall_loss}
        	\STATE Compute update with \eqref{eqn:param_upd}
        	\STATE $ m \leftarrow m+1 $ 
        	\ENDWHILE
        \STATE $ n \leftarrow n+1 $
        \ENDWHILE
        \end{algorithmic}
        \label{alg:offline}
        \end{algorithm}

        We split the set of sequences $ \{\textbf{X}_i\}_{i=1}^I $ and the sampling times $ \{\textbf{t}_i\}_{i=1}^I $ into three sets: the training $ \{\textbf{X}_i\}_{i=1}^{I_\text{train}} $; the validation $ \{\textbf{X}_i\}_{i=I_\text{train}}^{I_\text{validation}} $; the test set $ \{\textbf{X}_i\}_{i=I_\text{validation}}^{I} $. Using the training set and the weighted loss function, we train our model as described in Algorithm \ref{alg:offline}. We first initialize a random set of model parameters. Then, for each epoch, we iterate over mini-batches of sequences, which we sample randomly from the training set. For each mini-batch, we compute the loss function and finally compute the gradients of the loss function with respect to each model parameter. For the performance measurement on the validation set, we trace the loss on the validation set through epochs. If the validation loss stops decreasing for certain number of epochs, $ n_\text{tol} = 3 $, we stop the training of the model to prevent overfitting. We also choose the hyperparameters of our algorithm by comparing their validation losses after the training is finished. The set of hyperparameters consists of the parameters: $ p $, the state vector dimension; $ B $, the batch size; $ \tau $, the dimension of the nonlinear functions of the sampling time $ \mathbf{\Delta}_i^{(k)} $; $ \lambda $, the regularization parameter in the classifier loss; $ \alpha $, the loss weight. 
        
        After the training and the validation is finished, we finally compute the area under curve (AUC) score of the Receiver Operating Characteristics (ROC) curve on the test set \cite{bradley1997use}. In the ROC curves, we plot the True Positive Rate (TPR) and False Positive Rate (FPR) as a function of threshold. AUC score for the ROC curve is a well-known performance metric for the binary classification tasks \cite{bradley1997use}.

        \begin{algorithm}
        \caption{Online anomaly detection training algorithm.}
        \begin{algorithmic}
        \STATE Randomly initialize the LSTM model.
        \STATE Randomly initialize the One-Class Classifier.
        \WHILE{$ \textbf{X} $ arrives}
            \STATE $ H \leftarrow [] $, $ \hat{\textbf{X}} \leftarrow [] $ $ k \leftarrow 1 $
            \WHILE{$ k < K + 1 $}
                \STATE $ \textbf{h}^{(k)} \leftarrow \text{LSTM}(\textbf{x}^{(k)}) $
                \STATE $ H \leftarrow [H,\ \textbf{h}^{(k)}] $
                \STATE $ \hat{\textbf{X}} \leftarrow [\hat{\textbf{X}}, f^D(\textbf{h}^{(k)})] $
                \STATE $ k \leftarrow k+1 $
            \ENDWHILE
            \STATE $ \bar{\textbf{h}} \leftarrow \text{Pool}(H) $
            \STATE Compute $ \Lagr $ using \eqref{eqn:overall_loss}
            \STATE Compute parameter update \eqref{eqn:param_upd}
        \ENDWHILE
        \end{algorithmic}
        \label{alg:online}
        \end{algorithm}
        
        Although Algorithm \ref{alg:offline} requires the whole training set for the parameter optimization, we introduce an algorithm for the online setup where we update the model parameters with a single observed sequence. This procedure is given in Algorithm \ref{alg:online}. While a new sequence arrives, we compute the decision for this sequence using the most recent model parameters. After the decision is computed, we compute the loss function for the sequence and update the model parameters using the SGD. For the online setup, we choose the hyperparameters with the Particle Swarm Optimization (PSO) method where we train multiple models with different hyperparameters simultaneously \cite{lorenzo2017particle}. After a certain number of samples, $ n_\text{samples} = 100 $, are observed, we continue the training with the hyperparameter set that yields the least loss on the observed values so far.

        \begin{remark}
        	We optimize $ \alpha $ by considering only the reconstruction loss computed on the validation set. For the online setup, we instead compute the reconstruction loss on the observed samples. We pick the $ \alpha $ value that yields the least reconstruction loss on these sets.
        \end{remark}
        
        \begin{remark}
            Although we introduce the optimization procedures with the SGD updates, it is possible to use any other optimizer such as the ADAM optimizer \cite{kingma2014adam}. Moreover, we observe a faster convergence in our experiments compared to the SGD when we use the ADAM optimizer.
        \end{remark}
        
        In the following section, we describe the experimentation setup to test our model on real-life datasets.


    \begin{figure*}[h]
          \centering
          \subfigure[ROC curve for the A-LSTM, D-LSTM and M-LSTM model for 0.1 drop rate.]{\includegraphics[scale=0.329]{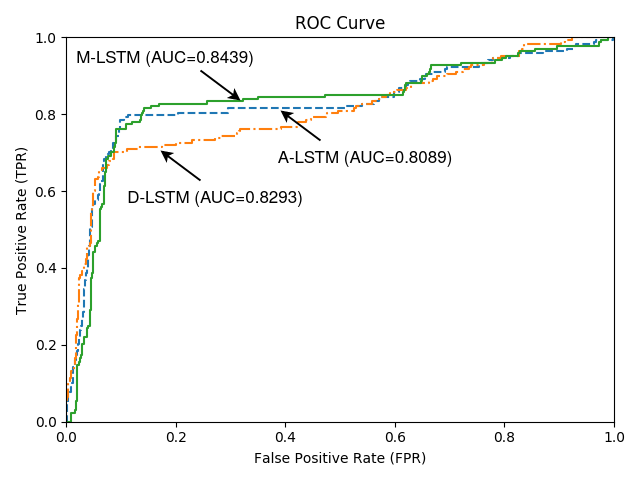}}\quad
          \subfigure[ROC curve for the A-LSTM, D-LSTM and M-LSTM model for 0.3 drop rate.]{\includegraphics[scale=0.329]{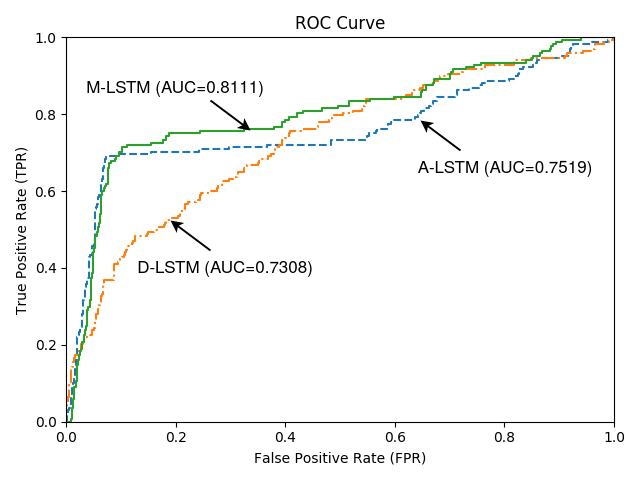}}
          \subfigure[ROC curve for the A-LSTM, D-LSTM and M-LSTM model for 0.5 drop rate.]{\includegraphics[scale=0.329]{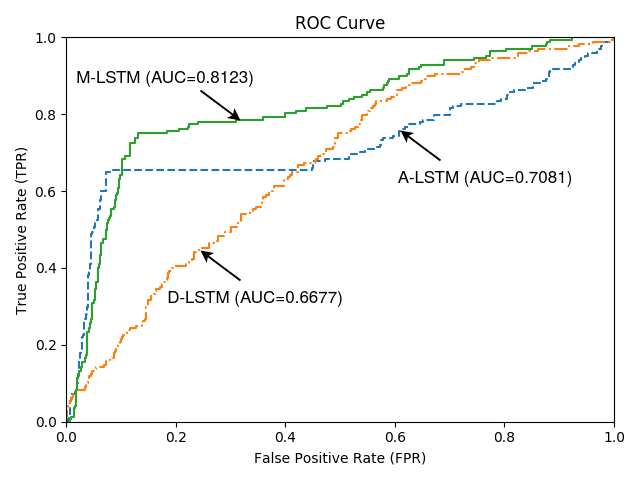}}\quad
          \subfigure[ROC curve for the A-LSTM, D-LSTM and M-LSTM model for 0.7 drop rate.]{\includegraphics[scale=0.329]{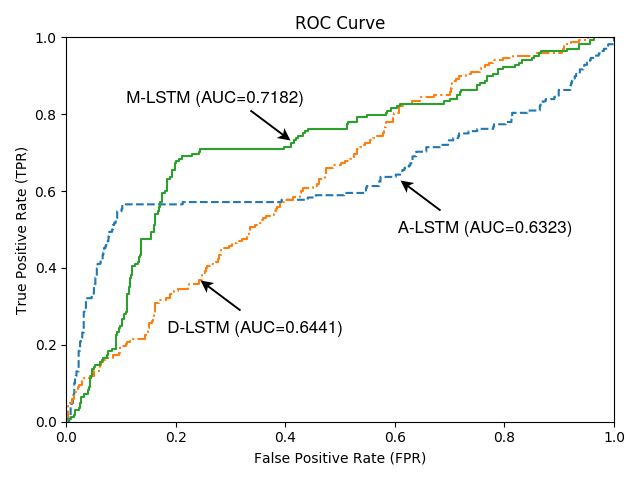}}
        \caption{ROC curves obtained for the three models, A-LSTM, D-LSTM and M-LSTM, on the validation set of the handwritten digit dataset. As it can be seen from the figures, AUC scores are relatively closer for small drop rates. Performance difference becomes more evident as the drop rate increases.}
        \label{fig:graphs_1}
    \end{figure*}

\section{Experiments}\label{sec:experiments}
    We first give a brief description of the real-life datasets and the experiment setup to measure the performance of our anomaly detection algorithm. We use two real-life datasets for performance measurements using the AUC score. Finally, we also compare the performances of different algorithms such as the LSTM with the sampling time incorporated as an additive term in the input, which we refer as A-LSTM. We also test the LSTM with the exponential decay mechanism, which we refer as D-LSTM and our LSTM model with time modulation gates, which we refer as M-LSTM.

    \subsection{Pen Based Handwritten Digit Recognition Dataset}
        In our first experiment, we use the pen digit trajectory dataset \cite{Dua2019}. This dataset consists of trajectories of handwritten digits written by 44 writers. Hence, the sequences consist of horizontal and vertical coordinates of the pen trajectory that is uniformly sampled with 100 ms periods. All trajectories have different number of elements due to different writers and digits. Instead of using the absolute coordinate values of the trajectories, we take the first order difference of the coordinate values and obtain the displacement vector at each element. After taking the first order difference, we randomly drop some of the elements from the trajectory sequences to simulate the missing values or irregular sampling in sequences. Finally, we convert the multi-class labels by grouping the classes $ \{1, 2, 4, 5, 7\} $ as the nominal class and the class $ 0 $ as the anomaly. Thus, we obtain a dataset with binary labels and negative class ratio of 0.87.
        
        \begin{table}[t]
            \begin{tabular}{|l|l|l|l|l|l|l|}
            \hline
            \textbf{Models} & B & p & $ \alpha $ & $ \lambda $ & $ \tau $ & $ \gamma $ \\ \hline
            A-LSTM	  & 32 & 8 & 1000 & 0.3 & - & -						\\ \hline
            D-LSTM    & 32 & 8 & 1000 & 0.3 & - & 0.1                        \\ \hline
            M-LSTM    & 32 & 8 & 1000 & 0.4 & 10 & -                        \\ \hline
            \end{tabular}
        \caption{Hyperparameters used for the models on the handwritten digit dataset.}
        \label{table:digit_params}
        \end{table}
        
        Using this dataset, we obtain the ROC curves shown in Fig. \ref{fig:graphs_1} for three different random drop probabilities, $ \{0.1, 0.3, 0.5, 0.7\} $. Moreover, the hyperparameters that we have selected for the models are shown in Table \ref{table:digit_params}. As seen from the Fig. \ref{fig:graphs_1}, we can see that the M-LSTM model significantly outperformed the other approaches. Although the performances are relatively closer for small drop rates, the difference gets larger as the random drop rate increases. This result shows that the modulation gates in the M-LSTM approach are more successful in modeling the irregular sampling or missing values in the sequences. Since we optimize the LSTM parameters with the reconstruction loss, we can conclude that the representations generated by the M-LSTM are more discriminative for anomaly detection compared to other approaches thanks to the modulation gates. Performance of the A-LSTM is worse compared to the other models as its reconstruction and modeling capacity is limited by the additive incorporation of the sampling time information, which is not as powerful as the D-LSTM or M-LSTM as they introduce scaling and modulation effects on the output.

    \begin{figure*}[h]
          \centering
          \subfigure[ROC curve for the A-LSTM, D-LSTM and M-LSTM, model for 0.1 drop rate.]{\includegraphics[scale=0.329]{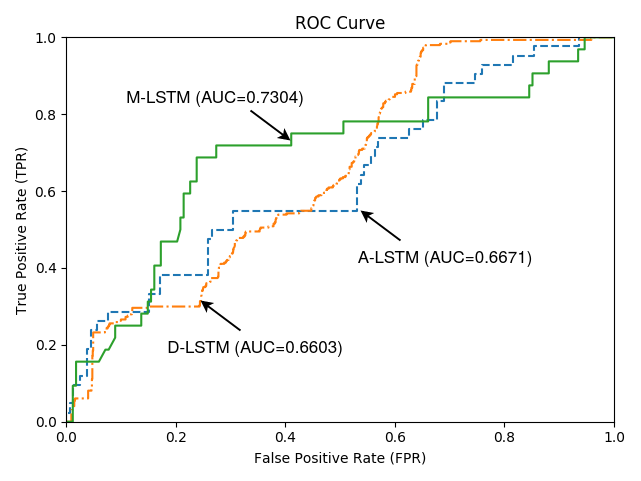}}\quad
          \subfigure[ROC curve for the A-LSTM, D-LSTM and M-LSTM, model for 0.3 drop rate.]{\includegraphics[scale=0.329]{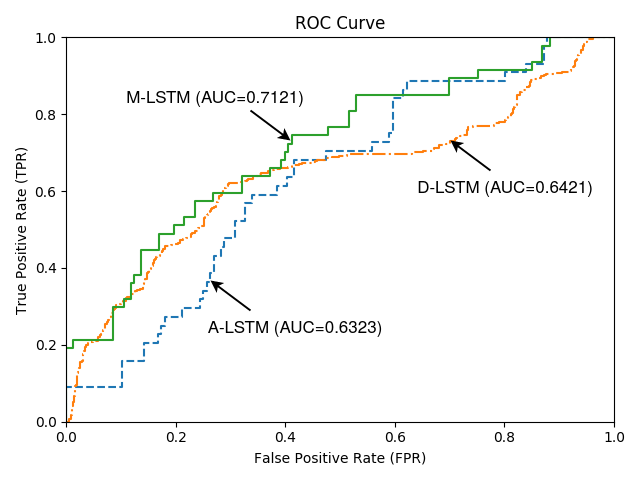}}
          \subfigure[ROC curve for the A-LSTM, D-LSTM and M-LSTM, model for 0.5 drop rate.]{\includegraphics[scale=0.329]{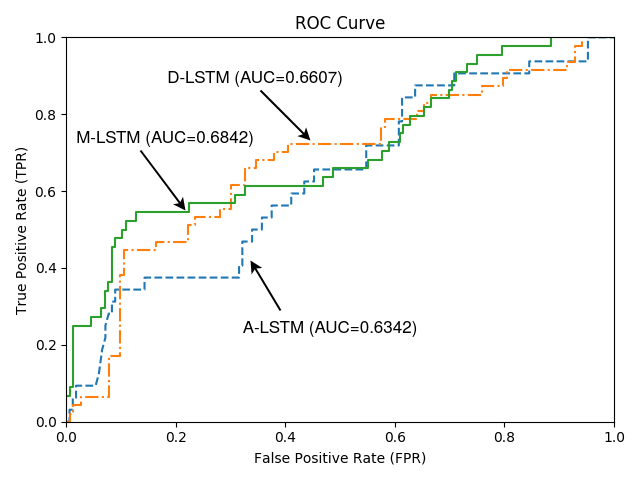}}\quad
          \subfigure[ROC curve for the A-LSTM, D-LSTM and M-LSTM, model for 0.7 drop rate.]{\includegraphics[scale=0.329]{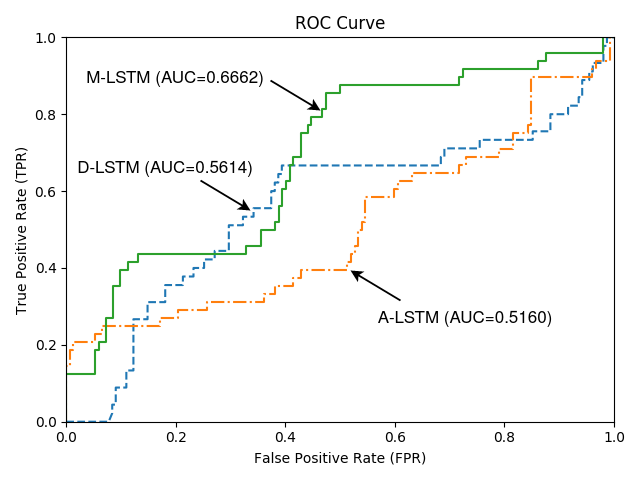}}
        \caption{ROC curves obtained for the two models, A-LSTM, D-LSTM and M-LSTM,, on the validation set of the activity recognition dataset. As it can be seen from the figures, AUC scores are relatively closer for small drop rates. Performance difference becomes more evident as the drop rate increases.}
        \label{fig:graphs_2}
        \end{figure*}
    
    \subsection{Daily and Sports Activities Data Set}
        In the second experiment, we use the sports and daily activity recognition dataset \cite{Barshan2019}. This dataset consists of motion sensor data of 19 daily and sports activities collected from 8 subjects for 5 minutes. Motion sensor data is collected from 5 body parts: torso, arms and legs, each having three accelerometers. In total there are 480 samples that are sampled with 40 ms period. All trajectories have the same length, however, different subjects perform activities in their own style. Again, we simulate the irregular sampling by randomly removing elements from the sequences to simulate the missing values or irregular sampling. Moreover, individual sequences have $ 60\times 125 $ samples. In order to decrease the memory and computation cost required during the temporal feature extraction, we randomly sample non-overlapping sequences with lengths between 55 and 75 (uniformly sampled). In total, we sample 2000 training samples, 1000 validation samples and 1000 test samples. Finally, we label the activities $ \{0, 1, 2, 3\} $ as anomaly and the rest as nominal. Anomalous activities correspond to sitting, standing, lying on back and on right side. In each of these sets, the negative class ratio is 0.9. 
        
        \begin{table}[]
            \begin{tabular}{|l|l|l|l|l|l|l|}
            \hline
            \textbf{Models} & B & p & $ \alpha $ & $ \lambda $ & $ \tau $  & $ \gamma $ \\ \hline
            A-LSTM	  & 32 & 16 & 1000 & 0.3 & - & -						\\ \hline
            D-LSTM     & 32 & 16 & 10000 & 0.3 & - & 0.1                        \\ \hline
            M-LSTM    & 32 & 32 & 1000 & 0.4 & 10 & -                        \\ \hline
            \end{tabular}
        \caption{Hyperparameters used for the models on the activity recognition dataset.}
        \label{table:activity_params}
        \end{table}
        
        On this dataset, we obtain the ROC curves as in Fig. \ref{fig:graphs_2}. Random dropping probabilities are the same as in the previous test setup. The hyperparameters are as in Table \ref{table:activity_params}. Similar to the previous experiment, we observe that the performances of the compared models are relatively close for small random drop probabilities. This is due to the ability of the LSTM models, which can model the sequences even if a portion of the sequence is missing. As the random dropping probability increases, the performance difference also increases. This is due to the time modulation mechanism, which allows the M-LSTM to model longer sampling distances. Even though the sequence lengths are relatively longer compared to the previous experiment, M-LSTM still significantly outperformed the other models. Finally, we can observe that the performance gap between the A-LSTM and the rest of the approaches is relatively larger in this setup. This is due to the continuous nature of the dataset as the M-LSTM can successfully model the sequences with the reconstruction loss under missing values or irregular sampling cases.

\section{Concluding Remarks}\label{sec:conclusion}
    In this paper, we introduce an unsupervised and online anomaly detection algorithm, which can detect anomalous sequences that are irregularly sampled or have missing values. Our approach is applicable to the sequences with variable lengths. In order to model irregularly sampled sequences, we use modulation gates in the standard LSTM structure and introduce a novel method to extract temporal features from variable length sequences. Later, we detect anomalies from the temporal features by estimating the support of the data via a one-class classifier, the SVDD. Here, all the parameters, e.g. the SVDD parameters and LSTM parameters are jointly optimized end-to-end where we introduce a combination of one-class classification loss and reconstruction loss. Finally, we extend our approach to the online setup with sequential gradient updates. Through extensive set of experiments on real-life datasets, we demonstrate that our algorithm outperforms the standard approaches thanks to the effective incorporation of the sampling time information to the algorithm via the modulation gates, using the reconstruction loss and end-to-end joint optimization, where we both optimize the features to be fed to the SVDD and the SVDD with respect to the optimized features.

\begin{spacing}{.87}
\bibliographystyle{IEEEtran}
\bibliography{main}

\begin{thebibliography}{10}
\providecommand{\url}[1]{#1}
\csname url@samestyle\endcsname
\providecommand{\newblock}{\relax}
\providecommand{\bibinfo}[2]{#2}
\providecommand{\BIBentrySTDinterwordspacing}{\spaceskip=0pt\relax}
\providecommand{\BIBentryALTinterwordstretchfactor}{4}
\providecommand{\BIBentryALTinterwordspacing}{\spaceskip=\fontdimen2\font plus
\BIBentryALTinterwordstretchfactor\fontdimen3\font minus
  \fontdimen4\font\relax}
\providecommand{\BIBforeignlanguage}[2]{{%
\expandafter\ifx\csname l@#1\endcsname\relax
\typeout{** WARNING: IEEEtran.bst: No hyphenation pattern has been}%
\typeout{** loaded for the language `#1'. Using the pattern for}%
\typeout{** the default language instead.}%
\else
\language=\csname l@#1\endcsname
\fi
#2}}
\providecommand{\BIBdecl}{\relax}
\BIBdecl

\bibitem{kiciman2005detecting}
E.~Kiciman and A.~Fox, ``Detecting application-level failures in
  component-based internet services,'' \emph{IEEE Transactions On Neural
  Networks}, vol.~16, no.~5, pp. 1027--1041, 2005.

\bibitem{8038008}
A.~{Dal Pozzolo}, G.~{Boracchi}, O.~{Caelen}, C.~{Alippi}, and G.~{Bontempi},
  ``Credit card fraud detection: A realistic modeling and a novel learning
  strategy,'' \emph{IEEE Transactions On Neural Networks And Learning Systems},
  vol.~29, no.~8, pp. 3784--3797, 2018.

\bibitem{li2013anomaly}
W.~Li, V.~Mahadevan, and N.~Vasconcelos, ``Anomaly detection and localization
  in crowded scenes,'' \emph{IEEE Transactions On Pattern Analysis And Machine
  Intelligence}, vol.~36, no.~1, pp. 18--32, 2013.

\bibitem{ryan1998intrusion}
J.~Ryan, M.-J. Lin, and R.~Miikkulainen, ``Intrusion detection with neural
  networks,'' in \emph{Advances In Neural Information Processing Systems},
  1998, pp. 943--949.

\bibitem{9064715}
L.~{Li}, J.~{Yan}, H.~{Wang}, and Y.~{Jin}, ``Anomaly detection of time series
  with smoothness-inducing sequential variational auto-encoder,'' \emph{IEEE
  Transactions On Neural Networks And Learning Systems}, pp. 1--15, 2020.

\bibitem{7880607}
Q.~{Chen}, R.~{Luley}, Q.~{Wu}, M.~{Bishop}, R.~W. {Linderman}, and Q.~{Qiu},
  ``Anrad: A neuromorphic anomaly detection framework for massive concurrent
  data streams,'' \emph{IEEE Transactions On Neural Networks And Learning
  Systems}, vol.~29, no.~5, pp. 1622--1636, 2018.

\bibitem{chandola2010anomaly}
V.~Chandola, A.~Banerjee, and V.~Kumar, ``Anomaly detection for discrete
  sequences: A survey,'' \emph{IEEE Transactions On Knowledge And Data
  Engineering}, vol.~24, no.~5, pp. 823--839, 2010.

\bibitem{aleroud2012contextual}
A.~AlEroud and G.~Karabatis, ``A contextual anomaly detection approach to
  discover zero-day attacks,'' in \emph{2012 International Conference On Cyber
  Security}.\hskip 1em plus 0.5em minus 0.4em\relax IEEE, 2012, pp. 40--45.

\bibitem{7412753}
D.~{Ienco}, R.~G. {Pensa}, and R.~{Meo}, ``A semisupervised approach to the
  detection and characterization of outliers in categorical data,'' \emph{IEEE
  Transactions On Neural Networks And Learning Systems}, vol.~28, no.~5, pp.
  1017--1029, 2017.

\bibitem{zhang2006anomaly}
J.~Zhang and M.~Zulkernine, ``Anomaly based network intrusion detection with
  unsupervised outlier detection,'' in \emph{2006 IEEE International Conference
  On Communications}, vol.~5.\hskip 1em plus 0.5em minus 0.4em\relax IEEE,
  2006, pp. 2388--2393.

\bibitem{8267488}
Z.~{Ghafoori}, S.~M. {Erfani}, S.~{Rajasegarar}, J.~C. {Bezdek},
  S.~{Karunasekera}, and C.~{Leckie}, ``Efficient unsupervised parameter
  estimation for one-class support vector machines,'' \emph{IEEE Transactions
  On Neural Networks and Learning Systems}, vol.~29, no.~10, pp. 5057--5070,
  2018.

\bibitem{li2016scalable}
S.~C.-X. Li and B.~M. Marlin, ``A scalable end-to-end gaussian process adapter
  for irregularly sampled time series classification,'' in \emph{Advances In
  Neural Information Processing Systems}, 2016, pp. 1804--1812.

\bibitem{lipton2015learning}
Z.~C. Lipton, D.~C. Kale, C.~Elkan, and R.~Wetzel, ``Learning to diagnose with
  lstm recurrent neural networks,'' \emph{ArXiv Preprint arXiv:1511.03677},
  2015.

\bibitem{scholkopf2001estimating}
B.~Sch{\"o}lkopf, J.~C. Platt, J.~Shawe-Taylor, A.~J. Smola, and R.~C.
  Williamson, ``Estimating the support of a high-dimensional distribution,''
  \emph{Neural Computation}, vol.~13, no.~7, pp. 1443--1471, 2001.

\bibitem{erfani2016high}
S.~M. Erfani, S.~Rajasegarar, S.~Karunasekera, and C.~Leckie,
  ``High-dimensional and large-scale anomaly detection using a linear one-class
  svm with deep learning,'' \emph{Pattern Recognition}, vol.~58, pp. 121--134,
  2016.

\bibitem{8052229}
N.~{Görnitz}, L.~A. {Lima}, K.~{Müller}, M.~{Kloft}, and S.~{Nakajima},
  ``Support vector data descriptions and $k$ -means clustering: One class?''
  \emph{IEEE Transactions on Neural Networks and Learning Systems}, vol.~29,
  no.~9, pp. 3994--4006, 2018.

\bibitem{wang2004anomaly}
Y.~Wang, J.~Wong, and A.~Miner, ``Anomaly intrusion detection using one class
  svm,'' in \emph{Proceedings From The Fifth Annual IEEE SMC Information
  Assurance Workshop, 2004.}\hskip 1em plus 0.5em minus 0.4em\relax IEEE, 2004,
  pp. 358--364.

\bibitem{zhang2007one}
R.~Zhang, S.~Zhang, S.~Muthuraman, and J.~Jiang, ``One class support vector
  machine for anomaly detection in the communication network performance
  data,'' in \emph{Proceedings Of The 5th Conference On Applied
  Electromagnetics, Wireless And Optical Communications}.\hskip 1em plus 0.5em
  minus 0.4em\relax Citeseer, 2007, pp. 31--37.

\bibitem{ma2003time}
J.~Ma and S.~Perkins, ``Time-series novelty detection using one-class support
  vector machines,'' in \emph{Proceedings Of The International Joint Conference
  On Neural Networks, 2003.}, vol.~3.\hskip 1em plus 0.5em minus 0.4em\relax
  IEEE, 2003, pp. 1741--1745.

\bibitem{liao2005clustering}
T.~W. Liao, ``Clustering of time series data—a survey,'' \emph{Pattern
  Recognition}, vol.~38, no.~11, pp. 1857--1874, 2005.

\bibitem{vio2000reconstruction}
R.~Vio, T.~Strohmer, and W.~Wamsteker, ``On the reconstruction of irregularly
  sampled time series,'' \emph{Publications Of The Astronomical Society Of The
  Pacific}, vol. 112, no. 767, p.~74, 2000.

\bibitem{goller1996learning}
C.~Goller and A.~Kuchler, ``Learning task-dependent distributed representations
  by backpropagation through structure,'' in \emph{Proceedings Of International
  Conference On Neural Networks (ICNN'96)}, vol.~1.\hskip 1em plus 0.5em minus
  0.4em\relax IEEE, 1996, pp. 347--352.

\bibitem{8438512}
J.~{Song}, Y.~{Guo}, L.~{Gao}, X.~{Li}, A.~{Hanjalic}, and H.~T. {Shen}, ``From
  deterministic to generative: Multimodal stochastic rnns for video
  captioning,'' \emph{IEEE Transactions On Neural Networks And Learning
  Systems}, vol.~30, no.~10, pp. 3047--3058, 2019.

\bibitem{dey2017gate}
R.~Dey and F.~M. Salemt, ``Gate-variants of gated recurrent unit (gru) neural
  networks,'' in \emph{2017 IEEE 60th International Midwest Symposium On
  Circuits And Systems (MWSCAS)}.\hskip 1em plus 0.5em minus 0.4em\relax IEEE,
  2017, pp. 1597--1600.

\bibitem{hochreiter1997long}
S.~Hochreiter and J.~Schmidhuber, ``Long short-term memory,'' \emph{Neural
  Computation}, vol.~9, no.~8, pp. 1735--1780, 1997.

\bibitem{mei2017neural}
H.~Mei and J.~M. Eisner, ``The neural hawkes process: A neurally
  self-modulating multivariate point process,'' in \emph{Advances In Neural
  Information Processing Systems}, 2017, pp. 6754--6764.

\bibitem{sahin2018nonuniformly}
S.~O. Sahin and S.~S. Kozat, ``Nonuniformly sampled data processing using lstm
  networks,'' \emph{IEEE Transactions On Neural Networks And Learning Systems},
  vol.~30, no.~5, pp. 1452--1461, 2018.

\bibitem{scholkopf2001kernel}
B.~Sch{\"o}lkopf, ``The kernel trick for distances,'' in \emph{Advances In
  Neural Information Processing Systems}, 2001, pp. 301--307.

\bibitem{graves2013speech}
A.~Graves, A.-r. Mohamed, and G.~Hinton, ``Speech recognition with deep
  recurrent neural networks,'' in \emph{2013 IEEE International Conference On
  Acoustics, Speech and Signal Processing}.\hskip 1em plus 0.5em minus
  0.4em\relax IEEE, 2013, pp. 6645--6649.

\bibitem{ergen2017unsupervised}
T.~{Ergen} and S.~S. {Kozat}, ``Unsupervised anomaly detection with lstm neural
  networks,'' \emph{IEEE Transactions on Neural Networks and Learning Systems},
  pp. 1--15, 2019.

\bibitem{neil2016phased}
D.~Neil, M.~Pfeiffer, and S.-C. Liu, ``Phased lstm: Accelerating recurrent
  network training for long or event-based sequences,'' in \emph{Advances In
  Neural Information Processing Systems}, 2016, pp. 3882--3890.

\bibitem{9059022}
M.~{Sabokrou}, M.~{Fathy}, G.~{Zhao}, and E.~{Adeli}, ``Deep end-to-end
  one-class classifier,'' \emph{IEEE Transactions On Neural Networks And
  Learning Systems}, pp. 1--10, 2020.

\bibitem{8825555}
P.~{Wu}, J.~{Liu}, and F.~{Shen}, ``A deep one-class neural network for
  anomalous event detection in complex scenes,'' \emph{IEEE Transactions On
  Neural Networks And Learning Systems}, pp. 1--14, 2019.

\bibitem{chen2018learning}
M.-Y. Chen, T.-C. Huang, Y.~Shu, C.-C. Chen, T.-C. Hsieh, and N.~Y. Yen,
  ``Learning the chinese sentence representation with lstm autoencoder,'' in
  \emph{Companion Proceedings Of The The Web Conference 2018}, 2018, pp.
  403--408.

\bibitem{bradley1997use}
A.~P. Bradley, ``The use of the area under the roc curve in the evaluation of
  machine learning algorithms,'' \emph{Pattern Recognition}, vol.~30, no.~7,
  pp. 1145--1159, 1997.

\bibitem{lorenzo2017particle}
P.~R. Lorenzo, J.~Nalepa, M.~Kawulok, L.~S. Ramos, and J.~R. Pastor, ``Particle
  swarm optimization for hyper-parameter selection in deep neural networks,''
  in \emph{Proceedings Of The Genetic And Evolutionary Computation Conference},
  2017, pp. 481--488.

\bibitem{kingma2014adam}
D.~P. Kingma and J.~Ba, ``Adam: A method for stochastic optimization,''
  \emph{ArXiv Preprint arXiv:1412.6980}, 2014.

\bibitem{Dua2019}
\BIBentryALTinterwordspacing
D.~Dua and C.~Graff, ``Uci machine learning repository,'' 2017. [Online].
  Available: \url{http://archive.ics.uci.edu/ml}
\BIBentrySTDinterwordspacing

\bibitem{Barshan2019}
\BIBentryALTinterwordspacing
K.~Altun and B.~Barshan, ``Human activity recognition using inertial/magnetic
  sensor units,'' 2010. [Online]. Available:
  \url{http://archive.ics.uci.edu/ml}
\BIBentrySTDinterwordspacing

\end{thebibliography}
\end{spacing}
\end{document}